\documentclass{article}

\usepackage{PRIMEarxiv}

\usepackage[utf8]{inputenc} 
\usepackage[T1]{fontenc}    
\usepackage{hyperref}       
\usepackage{url}            
\usepackage{booktabs}       
\usepackage{amsfonts}       
\usepackage{nicefrac}       
\usepackage{microtype}      
\usepackage{lipsum}
\usepackage{fancyhdr}       
\usepackage{graphicx}       
\graphicspath{{media/}}     

\usepackage{subcaption}
\usepackage{tabularx}
\usepackage{booktabs}
\usepackage{url}
\usepackage{enumitem}

\title{Enhancing Disinformation Detection with Explainable AI and Named Entity Replacement}

\author{
  Santiago Gonz{\'a}lez Silot \\
  SINAI research group \\
  Advanced Studies Center in ICT (CEATIC)\\
  Universidad de Jaén, Spain\\
  \texttt{sgs00034@red.ujaen.es} \\
  \And
  Andr{\'e}s Montoro-Montarroso \\
  Department of Technologies and Information Systems \\
  Universidad de Castilla-La Mancha, Spain \\
  \And
  Eugenio Mart{\'i}nez C{\'a}mara \\
  SINAI research group\\
  Advanced Studies Center in ICT (CEATIC)\\
  Universidad de Jaén, Spain \\
  \And
  Juan Gómez-Romero \\
  Department of Computer Science and Artificial Intelligence\\
  Universidad de Granada, Spain \\
}

\begin{document}
\maketitle

\begin{abstract}
The automatic detection of disinformation presents a significant challenge in the field of natural language processing. This task addresses a multifaceted societal and communication issue, which needs approaches that extend beyond the identification of general linguistic patterns through data-driven algorithms. In this research work, we hypothesise that text classification methods are not able to capture the nuances of disinformation and they often ground their decision in superfluous features. Hence, we apply a post-hoc explainability method (SHAP, SHapley Additive exPlanations) to identify spurious elements with high impact on the classification models. Our findings show that non-informative elements (e.g., URLs and emoticons) should be removed and named entities (e.g., Rwanda) should be pseudo-anonymized before training to avoid models' bias and increase their generalization capabilities. We evaluate this methodology with internal dataset and external dataset before and after applying extended data preprocessing and named entity replacement. The results show that our proposal enhances on average the performance of a disinformation classification method with external test data in 65.78\% without a significant decrease of the internal test performance.
\end{abstract}


\section{Introduction}
Disinformation is the deliberate creation and dissemination of falsehoods, hoaxes, and conspiracy theories with the intention of causing harm and manipulating public opinion \cite{wardle2017information}. This phenomenon is not new; it has evolved alongside humanity and has been amplified by advances in communication through the Internet and social networks. The proliferation of \textit{fake news} concerning people and events has significantly impacted democratic societies \cite{Ecker2024}, highlighting the urgent need for solutions to counter falsehoods, including the application of Artificial Intelligence (AI) \cite{montoro-montarroso_fighting_2023}.

Automatic disinformation detection (ADD) has primarily been addressed within the field of Natural Language Processing (NLP) with Machine Learning (ML) techniques \cite{perez-rosas2018,kelk2022}, as the written format remains prevalent despite the increasing use of multimedia formats (audio, image, video). However, ADD faces several shortcomings related to:

\begin{itemize}
    \item Data: Creating labeled datasets for training prediction models requires substantial effort and may introduce biases. Determining the truthfulness of information can be inherently challenging and culturally dependent.
    \item Machine Learning: ML methods based on pattern recognition tend to identify general behaviors but struggle with less frequent cases, which is typical for false content. This behavior can reinforce stereotypes and lead to discrimination against minorities that are underrepresented in datasets.
    \item Interpretability of neural networks: The outcomes of deep neural networks, the prevalent technique in ML, are difficult to understand by merely observing their parameters. Additionally, ADD models are heavily dependent on the specific language and phenomenon under analysis, limiting their adaptability to other contexts.
    \item Evaluation metrics: Most studies in this field focus solely on accuracy metrics, neglecting key factors and emphasizing the role of non-informative features. This focus may skew and distort the classification and make the system vulnerable to adversarial attacks.
\end{itemize}

This paper investigates the potential of explainable AI methods (XAI) \cite{arrieta2020explainable} in NLP-based ADD to identify spurious features, i.e., those that influence the output of prediction models but are not significant in the problem domain. Our research has two primary objectives: first, to mitigate the bias inherent in the training data, and second, to enhance the generalizability and robustness of ML methods for ADD.

To this end, we applied SHAP (SHapley Additive exPlanations) \cite{shap}, a post-hoc XAI method, to a paradigmatic example of a classification model for ADD, specifically a RoBERTa-based neural network \cite{liu2019roberta} trained with the Constraint AAAI 2022 dataset \cite{patwa2020fighting}. The results indicated that the model was focusing on surface-level features highly specific to the dataset, such as URLs and emoticons, and named entities that could be misleading to the ADD process, such as  Rwanda or Netflix. Accordingly, we proposed retraining the model after removing these surface-level spurious features and replacing named entities with their entity categories. Our hypothesis was that while the classification metrics of the new model might decrease compared to the initial one, it would exhibit greater generalizability.

The results after external validation confirmed this hypothesis. Testing both models with two additional ADD datasets, namely CTF \cite{paka2021cross} and COVID-19 Fake News \cite{tashtoush2022deep}, yielded an average improvement in the F1-score of 39.32\% for the \textit{debiased model}. To further validate the hypothesis, we repeated the process using all combinations of the three datasets (one for training, validation, and internal testing; the other two for external testing), confirming average improvements of 77.68\% (CTF as internal) and 80.34\% (COVID-19 Fake News as internal).

In summary, the main contributions of this paper are the following:

\begin{enumerate}
    \item We illustrate the use of a post-hoc XAI method for identifying spurious features in text classification for ADD.
    
    \item We provide evidence that performance metrics calculated over dataset splits do not reliably indicate the quality of models in ADD.
    
    \item We demonstrate improved model generalizability and reduced bias within the ADD context by applying an extended preprocessing and Named Entity Replacement (NER).
        
\end{enumerate}

The remaining sections of this paper are organized as follows. Section 2 covers the background of ADD with NLP and XAI methods. Section 3 details the proposed pipeline for identifying spurious features using SHAP. Section 4 describes the training, validation, and testing of the compared models. Section 5 discusses the results obtained. Finally, section 6 summarizes the main conclusions and outlines directions for future work.

\section{Background}
\label{sec:related_work}

This section sheds light on the target challenge of this paper, the use of explainability methods to enhance the generalization classification capacity of automatic disinformation detection systems. Accordingly, we succinctly present the most salient works related to ADD using NLP in section \ref{ss_rw_fn_detection}, XAI in section \ref{subsec:xai_methods}, and XAI in ADD in section \ref{subsec:xai_disinfo}.

\subsection{Automatic disinformation detection}\label{ss_rw_fn_detection}

The spread of false information with the intention of misleading is a long-established problem, which has been exacerbated by the communication facilities of the Internet and, in particular, social networks. In the literature, we find different approaches to face up the detection of misinformation. We focus in this paper on those based on natural language methods \cite{BONDIELLI201938}, and particularly in those grounded in linguistic features \cite{shu2017}. In this line, in \cite{rashkin2017} it is claimed that \textit{fake news} often contain certain words suggesting that the content is not trustworthy. Likewise, a wide range of linguistic features is also used in \cite{perez-rosas2018} to detect fake content from different domains, which is also evaluated on false information written in other languages in \cite{faustini2020fake}. The relevance of linguistic features to identify disinformation is  assessed in \cite{solopova2023} on the news related to the war between Russia and Ukraine. In \cite{chen2023}, pre-trained word embeddings are used to represent linguistic knowledge. Since the meaning of fake content may go beyond superficial linguistic features, in \cite{bonet2021} the discourse structure is leveraged to detect \textit{fake news} in the medical domain.

There are also contributions to disinformation detection based on recent deep learning technologies, namely transformers and language models. In \cite{samadi2021}, the authors fine-tuned a BERT language model with a convolution neural network on \textit{fake news} datasets from different domains. The high performance of pre-trained language models is also shown in \cite{baruah2020}. A challenge in the detection of fake content is the multilingual nature of the Internet; therefore, a multilingual transformer model is evaluated in \cite{guo2023_ieee} for the identification of false information written in Chinese and English. The current trend in the literature is using large language models (LLMs) to detect disinformative content, as studied in \cite{Hu_Sheng_Cao_Shi_Li_Wang_Qi_2024,su2024}.

This paper specifically aims to analyze the behavior and performance of classification models based on LLMs applied to ADD. A similar goal is studied in \cite{su2023}, in which the authors find that LLMs tend to classify news automatically generated by other LLMs as deceptive, and conversely, false information written by humans as trustworthy. While the previous contributions are mostly based on the empirical comparison of the performance of LLMs for ADD, our work applies explainable methods to provide insights about the reasons of failure and guidance for  improvement.




\subsection{Explainable Artificial Intelligence}
\label{subsec:xai_methods}
Today's machine learning models produce outputs that are difficult to interpret due to their complex structures, e.g., neural networks with intricate aggregation operations and activation functions. This lack of transparency makes it challenging for humans to understand or modify the models' behavior. The emerging field of explainable artificial intelligence (XAI) seeks to address this by providing justifications for model outputs to enable human verification and understanding.

There are two main categories of explainability techniques \cite{dwivedi_explainable_2023, carrillo_individual_2021}: (i) inherently transparent models such as decision trees or KNN, and (ii) post-hoc explainability techniques. In this paper we focus on post-hoc techniques since NLP commonly uses transformers which are complex neural networks not inherently explainable. The most commonly used methods in the literature for post-hoc explainability are LIME (Local Interpretable Model-Agnostic Explanations) \cite{ribeiro_why_2016}, SHAP (SHapley Additive exPlanations) \cite{shap}, Anchors \cite{ribeiro_anchors_2018}, Saliency Maps and Counterfactuals.

In particular, we use SHAP, which describes the behavior of the model measuring the influence that each characteristic has on the prediction of the model. SHAP has a mathematical basis in the game-theoretic Shapley value \cite{shapley1953value}, defined as the average marginal contribution of an instance of a feature among all possible coalitions. Accordingly, we use this value to measure the importance of a set of words in the out of a model and thus assess its importance.

\subsection{Automatic disinformation detection with Explainable Artificial Intelligence}
\label{subsec:xai_disinfo}
ADD is a task with a large social component and a great subjectivity. Therefore, it is important to use models yielding understandable decisions, in which biases can be detected and mitigated. In the literature we can find several use cases of explainability in disinformation detection.

On the intrinsic explainability side there are several architectures ---such as dEFEND (Explainable FakE News Detection) \cite{shu2019defend}, Graph-aware Co-Attention Network (GCAN) \cite{lu2004graph}, and xFake \cite{yang2019xfake}--- that use the attention weights of the news and external components (e.g., user comments) as an explainability mechanism. On the other hand, among the post-hoc methods LIME and SHAP are prevalent \cite{gongane2024survey}. Remarkably, in \cite{kozik2024explainability} it is shown that SHAP can be exploited by malicious users to design adversarial attacks by changing a selected number of words in a news item.

\section{A two-stage post-hoc XAI method to improve generalization in ADD}
\label{sec:methodology}

This section describes the methodology proposed in this paper. First, the methodology is shown together with the description of the different steps of which it is composed. Subsequently, the experiments carried out during the development of this work that have helped to develop the methodology are described, starting from a state-of-the-art ADD model and applying SHAP to observe what the model pays attention to.

\begin{figure*}[tpb]
    \centering
    \includegraphics[width=.8\linewidth]{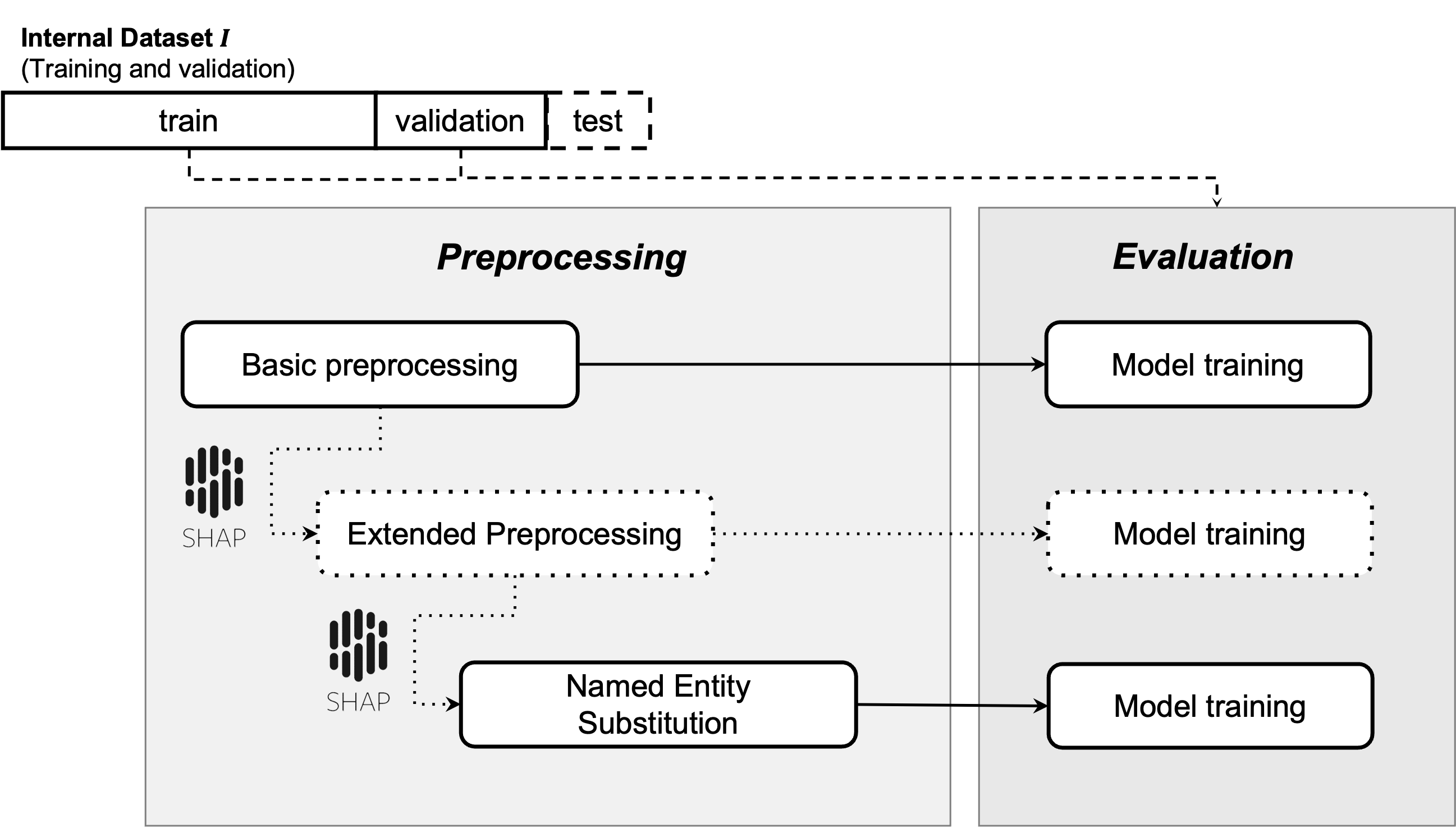}
    \caption{The outline of our methodology to enhance disinformation detection.}
    \label{fig:Workflow}
\end{figure*}

\begin{table}[tpb]
\centering
\small
\begin{tabularx}{0.45\textwidth}{l@{\phantom{0}}X@{\phantom{0}}l}
\toprule
Token & Meaning                                                  & Example      \\ \midrule
PER   & Person                                                   & Donald Trump \\
ORG   & Organization                                             & CDC          \\
LOC   & Location                                                 & California   \\
MISC  & Various entities (events, nationalities, products, etc.) & Spanish\\ \bottomrule
\end{tabularx}
\caption{Entities to be detected and replaced in the NER process.}
\label{tab:ner0}
\end{table}

In Figure \ref{fig:Workflow}, we present a comprehensive visualization of the methodology proposed in this paper. First of all, our methodology starts with a ADD model. This model has been carefully crafted and fine-tuned to deliver cutting-edge performance, aligning it with the state of the art in this domain.Then we apply a post-hoc explainability technique such as SHAP (SHapley Additive exPlanations) to know what the model pays attention to and detect possible errors and biases and enhace the transparency and interpretability of the results generated by our model. SHAP allows us to gain valuable insights into the inner workings of the model, shedding light on the specific features and attributes that the model prioritizes when making decisions. This process serves a dual purpose: not only does it improve the transparency of our model, but it also empowers us to identify potential errors and biases in its output.

Subsequently, as part of our ongoing commitment to refining the system, we take into account the valuable feedback provided by SHAP. In response to the insights gleaned from SHAP analysis, we undertake an iterative improvement process, focusing initially on data preprocessing. The primary objective here is to optimize the dataset to ensure that the model, during both its training and prediction phases, concentrates solely on the most relevant tokens within the text. This optimization includes several crucial steps, such as the removal of hashtags and their associated content, elimination of emoticons, cleansing of URLs, and the extraction of ''@'' symbols and their corresponding user mentions from the text.

These preprocessing adjustments are integral to our overarching strategy of mitigating issues identified through SHAP analysis. By eliminating extraneous elements from the data, we strive to provide the model with a cleaner, more streamlined input that enhances its overall performance and minimizes the risk of amplifying biases or inaccuracies.

Furthermore, to address the issue of biases, a Named Entity Recognition (NER) process is incorporated into the methodology. This process is instrumental in categorizing and replacing various entities within the dataset, such as locations, persons, organizations, and miscellaneous entities (e.g., events, nationalities, products, etc). The goal is to represent these entities as tokens that correspond to their respective categories, as detailed in Table \ref{tab:ner0}. This strategic replacement of entities not only allows the model to capture the broader linguistic patterns related to object classes and person references but also ensures that it remains agnostic to specific entities.

Consequently, this comprehensive approach yields several significant advantages. It promotes a more robust, less biased, and fairer model that is capable of making more generalizable predictions. By addressing the specific issues raised by SHAP, refining data preprocessing, and implementing NER-based entity categorization, we create a more resilient system that is not only at the forefront of disinformation detection but also upholds ethical considerations regarding bias and fairness. This multi-faceted methodology represents a substantial leap forward in the quest for reliable and accurate disinformation detection.

\subsection{Baseline}

In this section we present the experiments carried out during this work that helped in the development of the final methodology presented in the section \ref{sec:methodology}. Specifically, we decided to focus the experimentation on disinformation detection models for the COVID-19 domain. We started the experimentation using as a baseline a disinformation detection model that obtains state-of-the-art results in an important competition such as Constraint AAAI \cite{gonzalez2022filfa}. This dataset contains 10,700 tweets about COVID-19: 5,100 fake and 5,600 true. A more detailed description of the dataset is provided in section \ref{sec:datasets_used}.

A disinformation detection model was created based on RoBERTa \cite{liu2019roberta}, which obtained relevant results in the CONSTRAINT AAAI competition, achieving a fifth position with a Macro-F1 of 98.41\% and a difference of 2.89 points with the winner of this one. The model that obtained this result is a Fine-Tuning of the base model DigitalEpidemology-V2. It is a pre-trained model with 97 million \textit{tweets}, created by DigitalEpidemologyLab in collaboration with FISABio \cite{DigitalEpidemology}. It should also be noted that it was the one used in the \textit{Ensemble} that won the competition \cite{g2tmn} with 98.61\% of Macro-F1. Therefore during Fine-Tuning of this model, the \textit{tweets} were preprocessed as indicated by the authors, i.e.:

\begin{itemize}[noitemsep]
    \item Changing all words to lowercase.
    \item Replacing the URLs with the token \$URL\$.
    \item Replacing the hashtag by the token \$HASHTAG\$.
\end{itemize}

The Constraint AAAI dataset consists of $10,700$ \textit{tweets} in English of which $5,100$ are fake and $5,600$ are true. The true news are collected from Twitter and provides useful information about COVID-19. On the other hand, disinformation comes from Twitter, Facebook, and WhatsApp in addition to different fact-checking sites such as Politifact, NewsChecker, Boomlive, etc \cite{patwa2020fighting}.

\subsection{Application of SHAP to the base model.}

First, SHAP to this ADD model. For this purpose, the SHAP library will be used, which can be used to add a layer of explainability to any Machine Learning model. To do so, the SHAP values have to be generated using the Explainer of the library.

\begin{figure}[htbp]
\centering
\subcaptionbox{``the cdc currently reports 99031 deaths. in general the discrepancies in death counts between different sources are small and explicable. the death toll stands at roughly 100000 people today.''}{\label{fig:a}\includegraphics[width=1\linewidth]{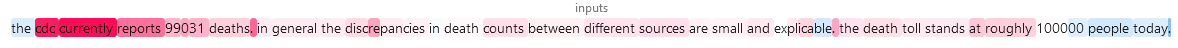}} \\
\subcaptionbox{``states reported 1121 deaths a small rise from last tuesday. southern states reported 640 of those deaths. \$URL\$''}{\label{fig:b}\includegraphics[width=1\linewidth]{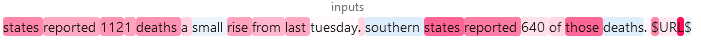}}
\caption{Two examples of sentences analyzed with SHAP. Words identified as more relevant to the ``true new'' category are highlighted in red, while those associated with the ``fake news'' category are highlighted in blue. The color intensity reflects the level of each word's importance to the model's prediction for that particular sentence.}
\label{fig:shap_example1}
\end{figure}

After obtaining the SHAP values for the complete dataset, we create visual representations of the results. In Figure \ref{fig:shap_example1}, we present sample sentences from the CONSTRAINT AAAI 2021 dataset. In these visualizations, words deemed more relevant to the ``true news'' class are highlighted in red, while those associated with the ``fake news'' class are highlighted in blue. The intensity of the color indicates the degree of relevance of each word to the model's prediction for that specific text.

\begin{figure}[htbp]
\centering
\subcaptionbox{Mean plot.}{\label{fig:a}\includegraphics[width=0.47\linewidth]{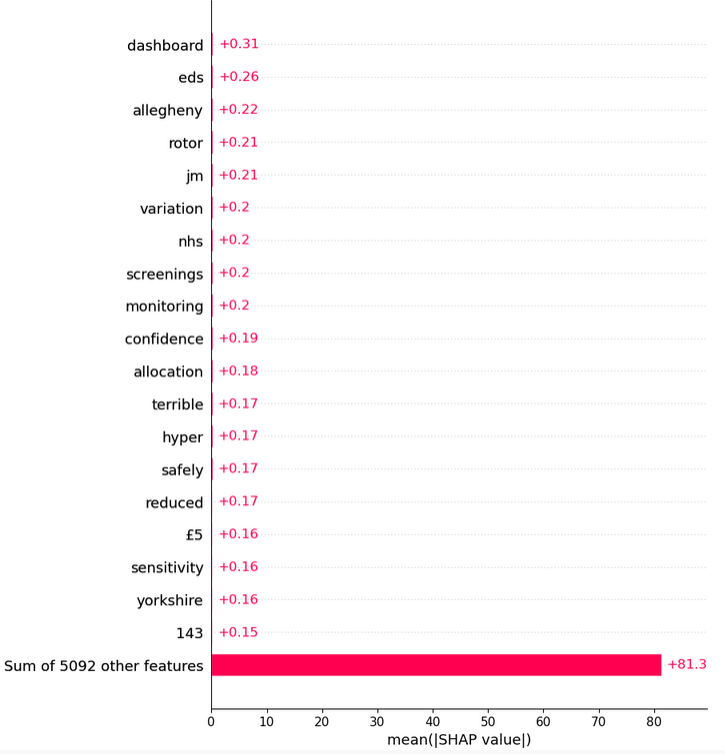}}
\subcaptionbox{Sum plot.}{\label{fig:b}\includegraphics[width=0.47\linewidth]{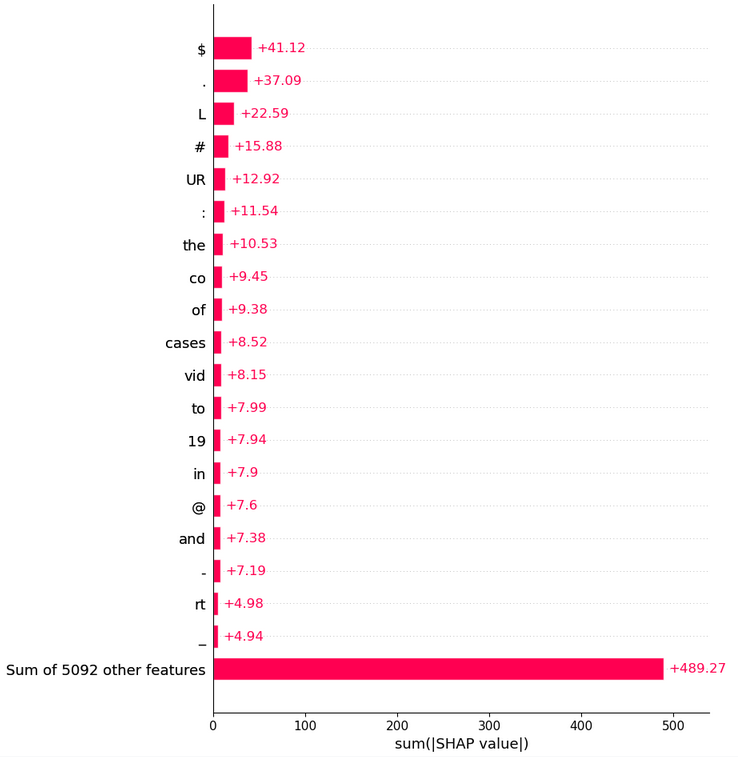}} \\
\caption{SHAP Global Bar plot, a global feature importance plot, where the global importance of each word is taken to be the mean (or sum) absolute value for that word over all the dataset.}
\label{fig:shap_global_bar_plot}
\end{figure}

As we can see in Figure \ref{fig:shap_global_bar_plot}, in some cases the model pays too much attention to certain spurious characters such as ``\$'' or the ``URL''. Also in some cases there are problems related to biases since the model uses information related to certain geographical entities (as in the case of ``southern'') or companies such as ``cdc''. This behavior of the models can raise ethical concerns, such as unfair treatment of certain entities, and can also introduce vulnerabilities, making them more susceptible to adversarial attacks.

In both cases, it does not make any sense for the model to use this type of tokens with such importance to discern the veracity of a news item. With this, we are beginning to see one of the hypotheses presented at the outset of this work. The model is not really learning to discern between true and false news but has learned a textual pattern (invaluable and illogical for human beings) present in the data set to differentiate between the 2 classes.

Part of the problem with this phenomenon is that this is not at all generalizable, if you were to evaluate the model with other datasets or put it into production and evaluate the performance of the model with texts that do not have these patterns, you would see how the performance probably drops. Overall, it seems that the model is closer to using misprocessed words, finding strange patterns and using biases related to named entities, rather than finding a pattern similar to what an expert would identify.

\subsection{Preprocessing improvement}
After all this analysis, it has been observed that on the one hand, the model does look at certain terms or words that seem coherent and are similar to those that a human would look at to discern between false and true news. On the other hand, it gives much value to meaningless terms, poorly processed words, spurious symbols, and meaningless patterns.

Therefore, to force the model to focus only on the really important tokens, it was decided to develop a preprocessing methodology that allows this. This methodology seeks to ensure that during the training and prediction phases, the model can only focus on relevant information about the text as an expert would do. Specifically, it was decided to apply the following preprocessing to the texts:

\begin{itemize}[noitemsep]
    \item Delete the \textit{hashtag} and its content.
    \item Remove emoticons.
    \item Remove URLS.
    \item Remove @ and their respective users.
\end{itemize}

Note that this preprocessing not only ensures that the model is set to what it should be but also avoids some bias by eliminating user-related information.

\begin{figure}[htbp]
\centering
\subcaptionbox{``A complex Sri Lankan herbal drink was said to remedy all virus infections''}{\label{fig:a}\includegraphics[width=1\linewidth]{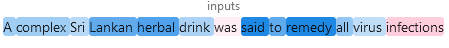}}\hspace{0.5pt}\\
\subcaptionbox{``Bolivia approved the use of chlorine dioxide amid the fight against covid-19.''}{\label{fig:b}\includegraphics[width=1\linewidth]{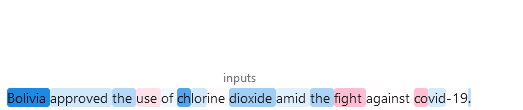}} 
\caption{Two examples of local explanations using SHAP after applying the methodology proposed in this paper. Words identified as more relevant to the ``true news'' category are highlighted in red, while those associated with the ``fake news'' category are highlighted in blue. The color intensity reflects the level of each word's importance to the model's prediction for that particular sentence.}
\label{fig:shap_newprepro}
\end{figure}

Once the new preprocessing is applied, the model is re-trained and the same process is applied again with SHAP to see how it has affected the model's learning and interpretability. As can be seen, the results have worsened a little with respect to the original model which has a Macro-F1 of 98.41\% and now has a 97.80\%. This is reasonable since this is since some of the patterns that the model used to discern are no longer there and although it makes the task more difficult, the objective is to look for more coherent patterns. To see how the model works after applying the methodology, a series of examples analyzed with SHAP can be seen in Figure \ref{fig:shap_newprepro}.

\begin{figure}[htbp]
\centering
{\label{fig:a}\includegraphics[width=0.7\linewidth]{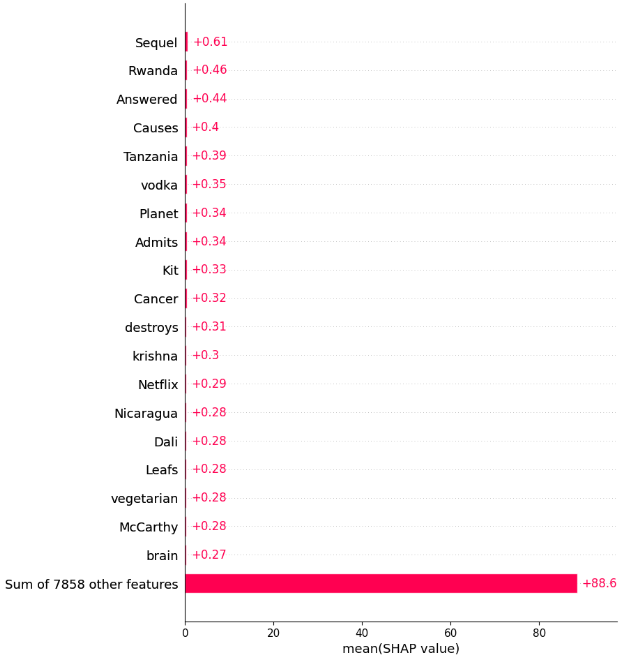}}
\caption{SHAP Global Bar plot (sum) after applying the methodology proposed in this paper.}
\label{fig:shap_newprepro_global_bar_plot}
\end{figure}

As can be seen in the examples in Figure \ref{fig:shap_newprepro}, by applying the new preprocessing, part of this bias and unwanted patterns have been eliminated since now only the relevant tokens in the text are fixed. Even so, as can be seen in the case of ``Sri Lanka'' or ``Bolivia'', a lot of weight is still being given to these terms that represent geographic entities and perhaps should not have so much relevance. Furthermore, in Figure \ref{fig:shap_newprepro_global_bar_plot}, you can see the terms that have the greatest impact on average. Although the words that appear now belong to the English dictionary and this list is not composed of spurious terms, they still do not, at first glance, determine that a news item is false or true. Additionally, there continue to be significant biases with certain entities such as ``Rwanda'', ``Tanzania'', ``Netflix'', or ``Nicaragua'' among others.

\subsection{Using Named Entity Recognition to avoid biases related to Entities.}
As we have seen throughout this work and specifically in last section, the ADD model has a serious problem with certain entities (countries, cities, people, and companies) because they rarely appear in the dataset and sometimes in a biased way (appearing mostly in one class). This is a major problem when dealing with the problem of misinformation from a fairness and ethics point of view.

It is for this reason that it has been decided to apply NER (Named Entity Recognition) to the dataset and replace each entity by a token representing its corresponding category (as can be seen in table \ref{tab:ner0}, in order to re-evaluate the model once it does not have the information about the entities in particular and only about its category (person, country, etc.). For this purpose, a BERT model trained for entity recognition (NER) will be used with the ``CoNLL-2003 Named Entity Recognition'' dataset \cite{ner}.


As explained in detail in the section \ref{sec:methodology}, once this process has been applied, the model is re-trained with the original parameters and the metrics on the test set are slightly worse, going from 98.41\% Macro-F1 to 96.34\%. However, this loss of performance is justified, since it allows us to obtain a more transparent model, safer and free of dangerous biases for society and because we also hypothesize that this step (and in general the whole methodology) increases the generalizability of the model in the ADD task on any similar dataset.

\section{Experimental configuration}
\label{sec:experiments}

\begin{figure*}[htbp]
\centering
{\label{fig:a}\includegraphics[width=1\linewidth]{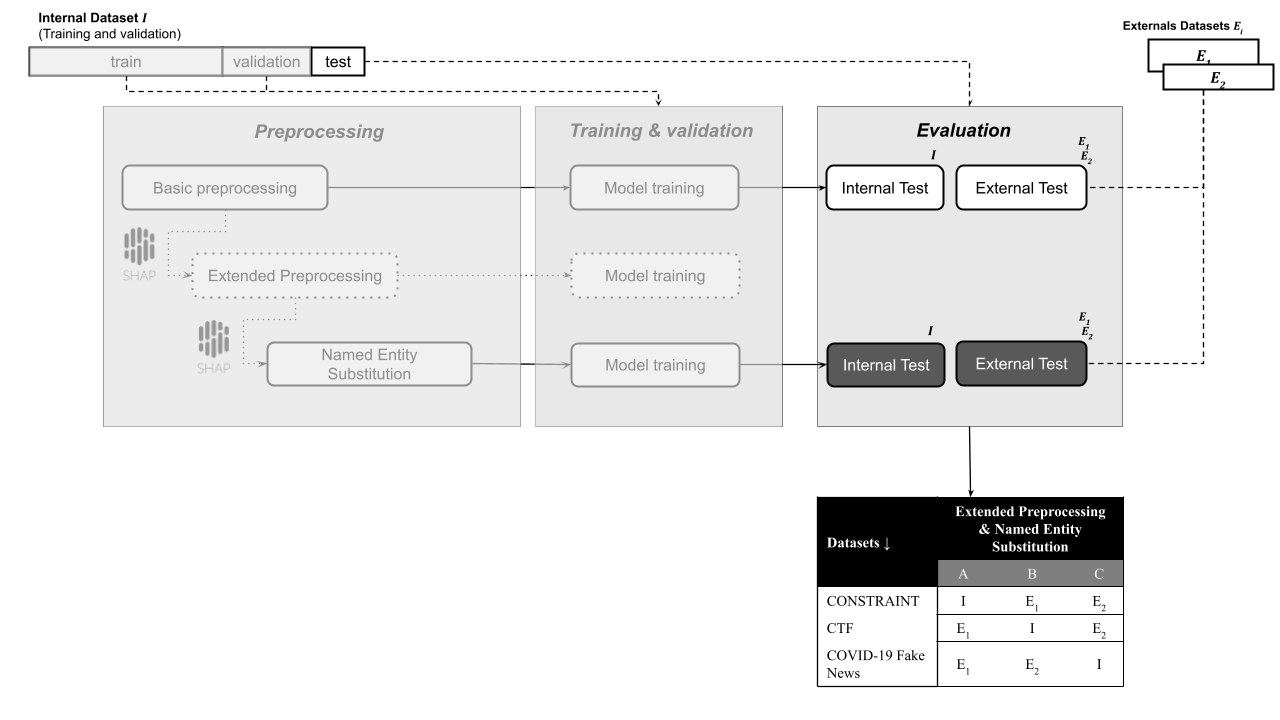}}
\caption{The outline of the experimental framework.}
\label{fig:experimental_conf}
\end{figure*}

We hypothesize that the metholodogy showed in section \ref{sec:methodology} serves to improve this type of disinformation detection systems in order to improve interpretability, absence of bias, and generalization capacity of our ADD model within the same context (disinformation detection about COVID-19 on social networks). In order to verify that this statement is true, it has been decided to apply an external evaluation methodology. That is, we will train the model with the internal dataset \textit{I} and evaluate the performance of the model for the dataset \textit{I} before and after applying the methodology. This will probably result in a slight loss of performance (compensated by all the improvements related with trustworthiness mentioned above). 

Subsequently the model trained with dataset \textit{I} will be re-evaluated before and after applying the methodology but not for the dataset \textit{I} for which it has been trained but for two similars (same genre and subject matter) but totally differents datasets \textit{$E_1$} and \textit{$E_2$}, with data which the original model has never seen. Thanks to this methodology, it will be possible to check the generalization capacity that this methodology brings to this type of models in disinformation detection.

In other words, we will take 3 datasets \textit{I}, \textit{$E_1$}, and \textit{$E_2$}. First, we will train the model with dataset \textit{I} and see the performance of the model for the dataset \textit{I} (internal dataset), \textit{$E_1$}, and \textit{$E_2$} (external validation datasets) before and after applying the proposed methodology. Subsequently, the external validation datasets (\textit{$E_1$}, and \textit{$E_2$}) will be exchanged with the training dataset \textit{I} and the process is repeated until all possible combinations are made, as can be seen in Figure \ref{fig:experimental_conf}.

\subsection{Datasets used}
\label{sec:datasets_used}
For the purpose just described, the following 3 datasets will be used:

\begin{enumerate}
    \item The \textbf{Constraint AAAI dataset} consists of $10,700$ \textit{tweets} in English of which $5,100$ are fake and $5,600$ are true. The true news are collected from Twitter and provides useful information about COVID-19. On the other hand, disinformation comes from Twitter, Facebook, and WhatsApp in addition to different fact-checking sites such as Politifact, NewsChecker, Boomlive, etc \cite{patwa2020fighting}.
    \item The \textbf{CTF} is the first labelled  COVID-19 misinformation dataset containing a total of 45.26K labelled tweets, with 18.55K labelled as 'genuine' and 26.71K as 'fake'. Genuine tweets in the dataset are primarly sourced from gobernmental health  or fact-checking websistes such as PolitiFact, Snopes, TruthOrFiction \cite{paka2021cross}.
    \item The \textbf{COVID-19 Fake News dataset} consist of 21.379 instances of which 11.365 are `real' and 10.014 are `fake' news related to COVID-19 which makes this dataset very well balanced. Data collection involves content from reputable sources like WHO, ICRC, UN, UNICEF and fact-checking websistes such as Snopes. PolitiFact and FactCheck. This dataset was originally cleaned by removing non English characters, numbers, hashtags, emojis, user mentions links and punctuation to standardise the data (which is very similar to the first step of the methodology proposed in this paper) \cite{tashtoush2022deep}.
\end{enumerate}

\begin{table*}[htbp]
\centering
\small
\caption{Comparation over the 3 datasets.}
\label{tab:comparativa_datasets}
\begin{tabularx}{\textwidth}{XXXXXX}
\hline
\textbf{Dataset} & \textbf{True News} & \textbf{Fake News} & \textbf{Total News} & \textbf{Source} & \textbf{Topic} \\ \hline
Constraint AAAI & 5,100 & 5,600 & 10,700 & Twitter & Covid-19  \\ \hline
CTF  & 18,550 & 26,710 & 45,260 & Twitter & Covid-19 \\ \hline
COVID-19 Fake News & 11,365 & 10,014 & 21,379  & Twitter & Covid-19\\ \hline
\end{tabularx}
\end{table*}

As can be seen in table \ref{tab:comparativa_datasets}, those three datasets have different data but from the same subject (disinformation about Covid-19) and from the same source (Twitter).

\section{Results and discussion}
\label{sec:discussion}

\begin{table*}[!t]
\centering
\small
\caption{Results using \textbf{Constraint} as \textbf{Train dataset}.}
\label{tab:resultados1}
\begin{tabularx}{\textwidth}{p{0.3\textwidth}XXX}
\hline
\textbf{Dataset} & \textbf{F1-Base model} & \textbf{F1-After methodology} & \textbf{Improvement} \\ \hline
Constraint AAAI (\textbf{internal}) & 98.50  & 96.95 & -1.57\% \\ \hline
CTF & 31.49 & 54.00 & 71.48\% \\ \hline
COVID-19 Fake News & 55.04 & 58.98 & 7.16\% \\ \hline
\multicolumn{2}{r}{Average Improvement in externals datasets:} & & \multicolumn{1}{X}{39,32} \\ \hline
\end{tabularx}
\end{table*}

\begin{table*}[!t]
\centering
\small
\caption{Results using \textbf{CTF} as \textbf{Train dataset}.}
\label{tab:resultados2}
\begin{tabularx}{\textwidth}{p{0.3\textwidth}XXX}
\hline
\textbf{Dataset} & \textbf{Base model} & \textbf{After methodology} & \textbf{Improvement} \\ \hline
CTF (\textbf{internal}) & 96.50 & 96.50 & 0\% \\ \hline
Constraint AAAI & 26.10 & 66.31 & 154.06\% \\ \hline
COVID-19 Fake News & 30.95 & 41.89 & 1.29\% \\ \hline
\multicolumn{2}{r}{Average Improvement in externals datasets:} & & \multicolumn{1}{X}{77,68\%} \\ \hline
\end{tabularx}
\end{table*}

\begin{table*}[!t]
\centering
\small
\caption{Results using \textbf{COVID-19 Fake News dataset} as \textbf{Train dataset}.}
\label{tab:resultados3}
\begin{tabularx}{\textwidth}{p{0.3\textwidth}XXX}
\hline
\textbf{Dataset} & \textbf{Base model} & \textbf{After methodology} & \textbf{Improvement} \\ \hline
COVID-19 Fake News (\textbf{internal}) & 99.14 & 90.29 & -8.92\% \\ \hline
CTF & 54.16 & 52.93 & -2.27\% \\ \hline
Constraint AAAI & 25.37 & 66.71 & 162.94\% \\ \hline
\multicolumn{2}{r}{Average Improvement in externals datasets:} & & \multicolumn{1}{X}{80,34\%} \\ \hline  
\end{tabularx}
\end{table*}

As we can see in Table \ref{tab:resultados1}, when we use Constraint AAAI as internal dataset and the other 2 as external validation, we see that Macro-F1 worsens slightly in the internal dataset after applying the methodology but has a large percentage of improvement in the external datasets, specifically around 24.97\% for CTF and 18\% for COVID-19 Fake News dataset, which gives us an average improvement of 21.49\% in the external datasets.

Table \ref{tab:resultados2} shows the same procedure, but in this case the CTF dataset is the training dataset and the other two are for external validation. In this case, surprisingly, there is a slight improvement in the internal dataset after applying the methodology, which may be due to the fact that the model is now more generalizable, it learns more useful and general characteristics of the domain to be solved, which makes it obtain better results in the test set than in the validation set. On the other hand, again we observe a large percentage of improvement in the external datasets, being 19.04\% in the CONSTRAINT AAAI dataset and 87.16\% in the COVID-19 Fake News dataset dataset. This significant improvement in the COVID-19 Fake News dataset dataset may be due to the fact that the dataset contains many biases, hashtags, URLS and other spurious features which have been treated in a better way after applying the methodology presented in this work. Lastly, we observe that on average we obtain a 53.10\% average improvement in the external datasets.

Finally in table \ref{tab:resultados3}, in which the training dataset is COVID-19 Fake News dataset, we can see that again in the training dataset there is a slight worsening after applying the methodology, even so again in the external dataset CONSTRAINT AAAI there is an improvement of 12.19\% but in this case in the CTF dataset there is a slight worsening of 3.8\%. Even so, the average improvement after applying the methodology is 4.15\%.

As it has been observed, by applying this evaluation method, the methodology developed in this work, not only makes the models less biased, fairer and paying attention to the really important tokens, but also increases the generalization capacity of the models in the disinformation detection task, which can be seen by the general improvement of the performance of the models in the external datasets, obtaining an average improvement after all the experiments of 65.78\% in Macro-F1.

\section{Conclusions}
\label{sec:conclusions}
This paper started with a disinformation classification model with high accuracy but a lack of explainability. Thanks to the use of an XAI technique, a preprocessing methodology has been developed to reduce model bias and increase generalization capabilities. The model has been evaluated after applying this methodology with two external datasets, achieving a global 65.78\% average improvement.

As future work we plan to explore three different areas of work. First, a custom language model specifically designed for the ADD domain could be created to improve explainability and handle domain-specific terms better. Secondly, the methodology developed for combating misinformation could be adapted to other NLP tasks, offering a broader application scope beyond its current domain. Finally, different explainability techniques, particularly Integrated Gradients, could be explored to gain deeper insights into the models' decision-making processes.

\section*{Acknowledgements}
\small
\noindent This publication is part of the projects XAI-DISINFODEMICS (PLEC2021-007681) funded by MICIU / AEI / 10.13039 / 501100011033 and by European Union NextGenerationEU / PRTR, FedDAP (PID2020-116118GA-I00) funded by MCIN / AEI / 10.13039 / 501100011033, and SAFER (PID2019-104735RB-C42) funded by MICIU / AEI/ 10.13039 / 501100011033.

\bibliographystyle{unsrt}  
\bibliography{references}  

\end{document}